\documentclass[10pt,twocolumn,letterpaper]{article}

\usepackage{cvpr}
\usepackage{times}
\usepackage{epsfig}
\usepackage{graphicx}
\usepackage{amsmath}
\usepackage{amssymb}
\usepackage[utf8]{inputenc} 
\usepackage[T1]{fontenc}    
\usepackage{url}            
\usepackage{booktabs}       
\usepackage{amsfonts}       
\usepackage{nicefrac}       
\usepackage{microtype}      
\usepackage{subcaption}
\usepackage{rotating}
\usepackage{appendix}

\cvprfinalcopy 

\def\cvprPaperID{49} 

\ifcvprfinal\fi
\begin{document}

\vspace{-3mm}
\title{Alleviating Semantic-level Shift:\\ 
A Semi-supervised Domain Adaptation Method for Semantic Segmentation}
\vspace{-3mm}
\author{Zhonghao Wang$^{1}$, Yunchao Wei$^{3}$, Rogerio Feris$^{2}$, 
Jinjun Xiong$^{2}$, \\Wen-mei Hwu$^{1}$,
Thomas S. Huang$^{1}$, Humphrey Shi$^{4,1}$\\
\\
{\small $^1$C3SR, UIUC, $^2$IBM Research, $^3$ReLER, UTS, $^4$University of Oregon}}
\vspace{-7mm}
\maketitle
\thispagestyle{empty}

\begin{abstract}
\vspace{-3mm}
   Utilizing synthetic data for semantic segmentation can significantly relieve human efforts in labelling pixel-level masks. A key challenge of this task is how to alleviate the data distribution discrepancy between the source and target domains, \emph{i.e.} reducing domain shift. The common approach to this problem is to minimize the discrepancy between feature distributions from different domains through adversarial training. However, directly aligning the feature distribution globally cannot guarantee consistency from a local view (\emph{i.e.} semantic-level). To tackle this issue, we propose a semi-supervised approach named Alleviating Semantic-level Shift (ASS), which can promote the distribution consistency from both global and local views. We apply our ASS to two domain adaptation tasks, from GTA5 to Cityscapes and from Synthia to Cityscapes. Extensive experiments demonstrate that: (1) ASS can significantly outperform the current unsupervised state-of-the-arts by employing a small number of annotated samples from the target domain; (2) ASS can beat the oracle model trained on the whole target dataset by over 3 points by augmenting the synthetic source data with annotated samples from the target domain without suffering from the prevalent problem of overfitting to the source domain.
\end{abstract}
\vspace{-3mm}

\section{Introduction}
Due to the development and use of deep learning techniques, major progress has been made in semantic segmentation, one of the most crucial computer vision tasks \cite{deeplabv1,deeplabv2,pspnet,deeplabv3,cheng2019spgnet,jiao2019geometry,huang2020alignseg}. However, the current advanced algorithms are often data hungry and require a large amount of pixel-level masks to learn reliable segmentation models. Therefore, one problem arises -- \emph{annotating pixel-level masks is costly in terms of both time and money.} For example, Cityscapes \cite{Cityscapes}, a real footage dataset, requires over 7,500 hours of human labor on annotating the semantic segmentation ground truth. 

\begin{figure}[t]
\centering
\includegraphics[width=.5\textwidth]{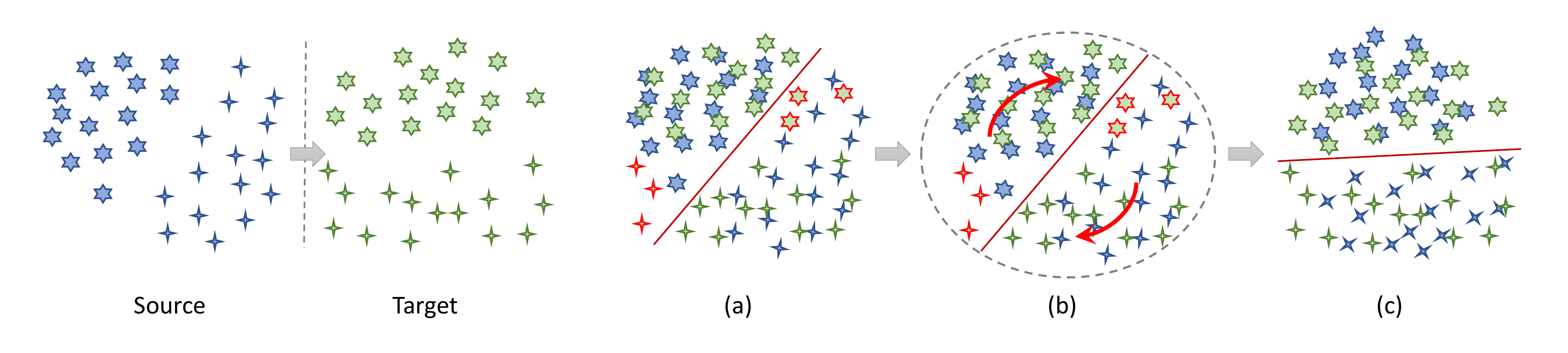}
\caption{Domain adaptation. (a) Global adaptation. (b) Semantic-level adaptation. (c) Ideal result.}
\label{figure1}
\vspace{-5mm}
\end{figure}

To tackle this issue, unsupervised training methods \cite{road, dagan, outputspace, fullyconv, wang2020differential} were proposed to alleviate the burdensome annotating work. Specifically, images labeled from other similar datasets (source domain) can be utilized to train a model and adapted to the target domain by addressing the domain shift issue. For the semantic segmentation task on Cityscapes dataset specifically, previous works \cite{gta5, synthia} have created synthetic datasets which cost little human effort to serve as the source datasets. 

\begin{figure*}
\centering
\includegraphics[width=1\textwidth]{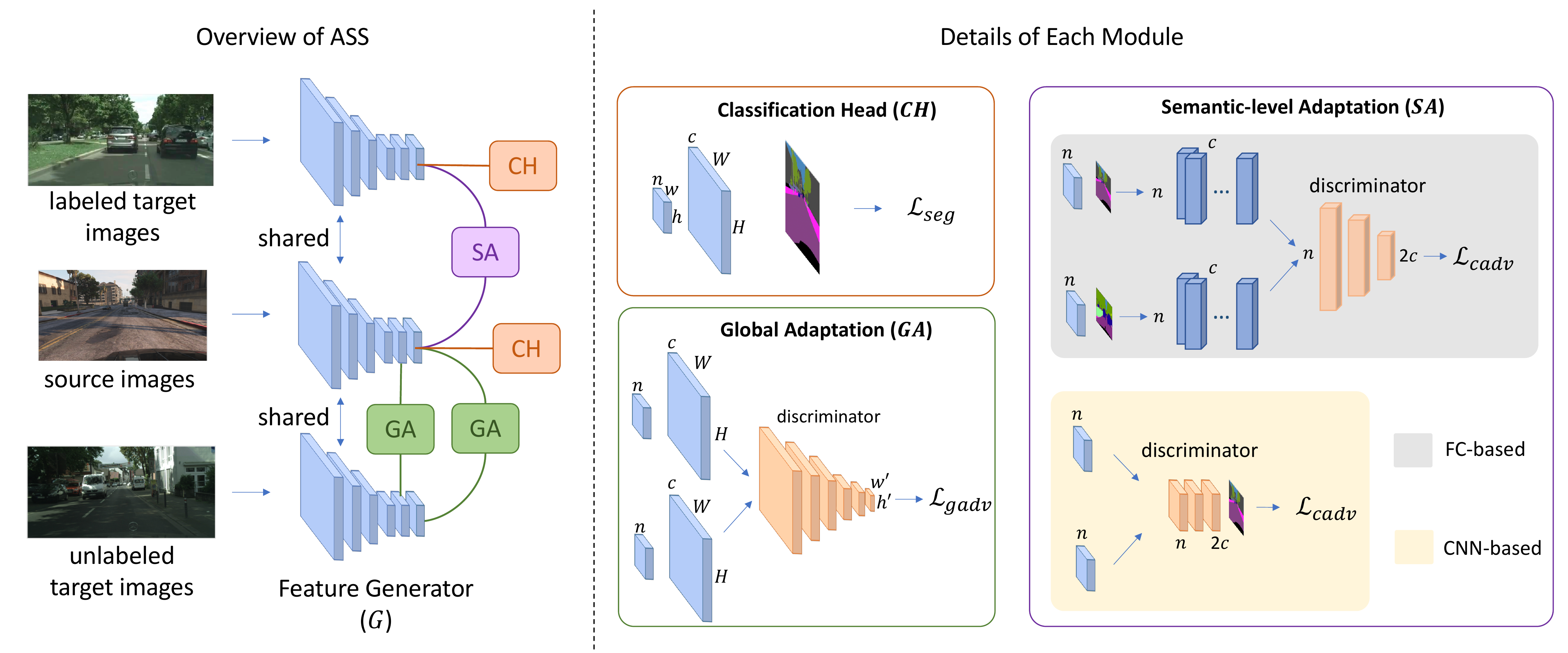}
\caption{Structure overview. $c$ is the number of classes for adaptation. $W$ and $H$ are the width and height of the input image respectively. $n$ is the number of feature channels of the feature map.}
\label{figure2}
\vspace{-6mm}
\end{figure*}

While evaluating the previous unsupervised or weakly-supervised methods for semantic segmentation \cite{outputspace, revisit, stc, zilong, selferasing, mining,qian2019weakly}, we found that there is still a large performance gap between these solutions and their fully-supervised counterparts. By delving into the unsupervised methods, we observe that the semantic-level features are weakly supervised in the adaptation process and the adversarial learning is only applied on the global feature representations. However, simply aligning the features distribution from global view cannot guarantee consistency in local view, as show in Figure \ref{figure1} (a), which leads to poor segmentation performance on the target domain. To address this problem, we propose a semi-supervised learning framework -- Alleviating Semantic-level Shift (ASS) model -- for better promoting the distribution consistency of features from two domains. In particular,  ASS not only adapts global features between two domains but also leverages a few labeled images from the target domain to supervise the segmentation task and the semantic-level feature adaptation task. In this way, the model can ease the inter-class confusion problem during the adaptation process (as shown in Figure \ref{figure1} (b)) and ultimately alleviate the domain shift from local view (as shown in Figure \ref{figure1} (c)). As a result, our method 1) is much better than the current state-of-the-art unsupervised methods by using a very small amount of the labeled target domain images; 2) addresses the prevalent problem that semi-supervised models typically overfit to the source domain \cite{survey}, and outperforms the oracle model trained with the whole target domain dataset by utilizing the synthetic source dataset and labeled images from the target domain. 

\vspace{-3mm}
\section{Related Works}
\vspace{-2mm}
\textbf{Semantic segmentation.} This task requires segment the pixels of images into semantic classes. Deeplab \cite{deeplabv1, deeplabv2, deeplabv3} is such a series of deep learning models that attained top on the 2017 Pascal VOC \cite{pascal} semantic segmentation challenge. It uses Atrous Spatial Pyramid Pooling (ASPP) module which combines multi-rate atrous convolutions and the global pooling technique to enlarge the field of view on the feature map and therefore deepen the model's understanding of the global semantic context. Deeplab v2 has laconic structure and good performance in extracting images features and can be easily trained. Therefore, it is selected as the backbone network for our work.

\textbf{Domain adaptation.} This task requires transfer and apply the useful knowledge of the model trained on the off-the-shelf dataset to the target task dataset \cite{domainadapt}. A typical structure for the domain adaptation is Generative Adversarial Networks (GAN) \cite{gan}. It consists of a discriminator that distinguishes which domain the input feature maps are from, and a generator that generates the feature maps to fool the discriminator. The discriminator thereby supervises the generator to minimize the discrepancy of the feature representations from the two domains.

\vspace{-1mm}
\section{Method: Alleviating Semantic-level Shift}
\vspace{-1mm}
We randomly select a subset of images from the target domain with ground truth annotations, and denote this set of images as \{$\mathcal{I_{T_L}}$\}. We denote the whole set of source images and the set of unlabeled target images as \{$\mathcal{I_S}$\} and \{$\mathcal{I_{T_U}}$\} respectively. As shown in Figure \ref{figure2}, our domain adaptation structure has four modules: the feature generation module $G$, the segmentation classification module $CH$, the global feature adaptation module $GA$ and the semantic-level adaptation module $SA$. We denote the output feature maps of $G$ by $F$, the ground truth label maps by $Y$ and the downsampled label maps (of the same height and width as $F$) as $y$. We use $H$, $W$ to denote the height and width of the input image, $h$, $w$ to denote those of $F$, and $h'$, $w'$ to denote those of the confidence maps output by the discriminator of $GA$. $C$ is the class set, $c$ is the number of classes, and $n$ is the channel number of $F$. When testing the model, we forward the input image to $G$ and use $CH$ to operate on $F$ to predict the semantic class that each pixel belongs to. The following sections will introduce the details of each module.

\vspace{-2mm}
\subsection{Segmentation}
We forward $F$ to a convolutional layer to output the score maps with $c$ channels. Then, we use a bilinear interpolation to upsample the score maps to the original input image size and apply a softmax operation channel-wisely to get score maps $P$. The segmentation loss $L_{seg}$ is calculated as
\begin{equation}
    L_{seg}(I) = -\sum_{H,W} \sum_{k\in C} Y^{(H,W,k)}\log (P^{(H,W,k)})
    \label{eq1}
\end{equation}
\vspace{-5mm}
\subsection{Global Feature Adaptation Module}
\label{3.2}
This module adapts $F$ from the source domain to the target domain. we input the source image score maps $P_s$ to the discriminator $D_g$ of $GA$ to conduct the adversarial training. We define the adversarial loss as:
\begin{equation}
    L_{gadv}(I_s) = -\sum_{h',w'}  \log (D_g(P_s)^{(h',w',1)})
    \label{eq2}
\end{equation}
We define $0$ as the source domain pixel and $1$ as the target domain pixel for the output of $D_g$. Therefore, this loss will force $G$ to generate features closer to the target domain globally. To train $D_g$, we forward $P_s$ and $P_{t_u}$ to $D_g$ in sequence. The loss of $D_g$ is calculated as: 
\begin{equation}
\begin{split}
    L_{gd}(P) = -\sum_{h',w'}  ((1-z)\log (D_g(P_s)^{(h',w',0)}) \\
    +z\log (D_g(P_t)^{(h',w',1)}))
\end{split}
\end{equation}
where $z=0$ if the feature maps are from the source domain and $z=1$ if the feature maps are from the target domain. 


\subsection{Semantic-level Adaptation Module}
This module adapts the feature representation for each class in the source domain to the corresponding class feature representation in the target domain to alleviate the domain shift from semantic-level. 
\vspace{-2mm}
\subsubsection{Fully connected semantic adaptation (FCSA)}
We believe that the feature representation for a specific class at each pixel should be close to each other. Thereby, we can average these feature vectors across the height and width to represent the semantic-level feature distribution, and adapt the averaged feature vectors to minimize the distribution discrepancy between two domains. The semantic-level feature vector $V_k$ of class $k$ is calculated as 
\begin{equation}
    V^{k} = \frac{\sum_{h,w} y^{(h,w,k)} F^{(h,w,:)}} {\sum_{h,w} y^{(h,w,k)}}
\end{equation}
where $k \in C$, $V^{k} \in \mathbb{R}^{n}$. Then we forward these semantic-level feature vectors to the semantic-level feature discriminator $D_s$ for adaptation, as shown in Figure~\ref{figure2}. $D_s$ only has 2 fully connected layers, and outputs a vector of $2c$ channels after a softmax operation. The first half and the last half channels correspond to classes from the source domain and the target domain respectively. Therefore, the adversarial loss can be calculated as 
\begin{equation}
    L_{sadv}(I_s) = -\sum_{k \in C}  \log (D_s(V_s^k)^{(k+c)})
\end{equation}
To train $D_s$, we let it classify the semantic-level feature vector to the correct class and domain. The loss of $D_s$ can be calculated as:
\begin{equation}
\begin{split}
    L_{sd}(V) = -\sum_{k \in C}  ((1-z)\log (D_s(V^k)^{(k)}) \\
    + z \log (D_s(V^k)^{(k+c)}))
\end{split}
\end{equation}
where $z=0$ if the feature vector is from the source domain and $z=1$ if it is from the target domain. 
\subsubsection{CNN semantic adaptation (CSA)}
We observe that it is hard to extract the semantic-level feature vectors, because we have to use the label maps to filter pixel locations and generate the vectors in sequence. Therefore, inspired from the previous design, we come up with a laconic CNN semantic-level feature adaptation module. The discriminator uses convolution layers with kernel size $1\times1$, which acts as using the fully connected discriminator to operate on each pixel of $F$, as shown in Figure~\ref{figure2}. The output has $2c$ channels after a softmax operation where the first half and the last half correspond to the source domain and the target domain respectively. Then, the adversarial loss can be calculated as:
\begin{equation}
    L_{sadv}(I_s) = -\sum_{h,w}  \log (D_s(F_s)^{(h, w, k+c)})
\end{equation}
where $k$ is the pixel ground truth class. To train the discriminator, we can use the loss as follows:
\begin{equation}
\begin{split}
    L_{sd}(F) = -\sum_{h,w}  ((1-z)\log (D_s(F)^{(h,w,k)})  \\
    + z \log (D_s(F)^{(h,w,k+c)}))
\end{split}
\end{equation}
\subsection{Adversarial Learning Procedure}
Our ultimate goal for $G$ is to have a good semantic segmentation ability by adapting features from the source domain to the target domain. Therefore, the training objective for $G$ can derive from Eqn (\ref{eq1}) as
\begin{equation}
\begin{split}
    L(I_s, I_{t_l})=\lambda_{seg}(L_{seg}(I_s)+L_{seg}(I_{t_l})) \\
    + \lambda_{gadv}L_{gadv}(I_s)+\lambda_{sadv}L_{sadv}(I_s)
\end{split}
\end{equation}
where $\lambda$ is the weight parameter. The two discriminators should be able to distinguish which domain the feature maps are from, which enables the features to be adapted in the right direction. We can simply sum up the two discriminator losses as the training objective for discriminative modules. 
\begin{equation}
\begin{split}
    L(F_s, F_{t_u}, F_{t_l})=\lambda_{gd}(L_{gd}(F_s)+L_{gd}(F_{t_u})) \\
    +\lambda_{sd}(L_{sd}(F_s)+L_{sd}(F_{t_l}))
\end{split}
\end{equation}
In summary, we will optimize the following min-max criterion to let our model perform better in segmentation task by adapting the features extracted from the source domain more alike the ones extracted from the target domain.
\begin{equation}
    \max_{D_g,D_s}\min_{G} L(I_s, I_{t_l})-L(F_s, F_{t_u}, F_{t_l})
    \label{opt_eqn}
\end{equation}

\section{Implementation}
\subsection{Network Architecture}
We follow \cite{outputspace} to build the network structures for the backbone network, the classification module (CH) and the global adaptation module (GA). For $FCSA$, we use two fully connected layers with channel number of 1024 and put a Leaky ReLU \cite{leakyrelu} of 0.2 negative slope between them, and twice the class number for the output. For $CSA$, we use two convolutional layers with the kernel size of 1$\times$1, stride of 1 and channel number of 1024 and twice the class number for the output. We insert a Leaky ReLU \cite{leakyrelu} layer with 0.2 negative slope between the two convolutional layers.

\vspace{-2mm}
\subsection{Network Training}
\vspace{-1mm}
We optimize Eqn (\ref{opt_eqn}) in an adversarial strategy. We first . We use Stochastic Gradient Descent (SGD) with Nesterov's method \cite{Nesterov} with momentum 0.9 and weight decay $5\times10^{-4}$ to optimize the segmentation network . Following \cite{deeplabv1}, we set the initial learning rate to be $2.5\times 10^{-4}$ and let it polynomially decay with the power of 0.9. We use Adam optimizer \cite{adam} with momentum 0.9 and 0.99 for all the discriminator networks. We set the initial learning rate to be $10^{-4}$ and follow the same polynomial decay rule. 
\vspace{-2mm}

\begin{table}[t]
\footnotesize
  \caption{GTA5 $\rightarrow$ Cityscapes: performance contributions of adaptation modules. The oracle model is only trained with the given number of Cityscapes labeled images. }
  \vspace{-2mm}
  \label{GTAtable}
  \centering
  \begin{tabular}{c|ccccc}
    \toprule
    \# City     & Oracle      & GA    & GA+FCSA   & GA+CSA    & Improve\\
    \hline
    0                     & -           & 42.4     & -             & -             & -\\
    50                    & 39.5       & 50.0     & 50.2         & 50.1         & +10.6\\
    100                   & 43.6       & 53.5     & 54.1         & 54.2         & +10.6\\
    200                   & 47.1       & 54.4     & 56.4         & 56.0         & +8.9\\
    500                   & 53.6       & 56.5     & 59.9         & 60.2         & +6.6\\
    1000                  & 58.6       & 58.0     & 63.8         & 64.5         & +5.9\\
    2975 (all)                 & 65.9       & 59.71     & 68.8         & 69.1         & +3.2\\
    \bottomrule
  \end{tabular}
  \label{gta_abl}
  \vspace{-2mm}
\end{table}
\begin{table}
  \caption{parameters analysis}
  \vspace{-2mm}
  \footnotesize
  \begin{subtable}{.49\linewidth}
  \centering
  \begin{tabular}{c|p{1.33cm}p{1.33cm}p{1.33cm}p{1.33cm}}
    \toprule
    \# City    & $\lambda=1$      & $\lambda=0.2$    & $\lambda=0.04$ & $\lambda=0.008$ \\
    \hline
    100                   & \textbf{54.11}       & 53.87     & 53.68     & 53.96     \\
    500                   & 59.76       & 59.29     & \textbf{59.89}     & 59.74     \\
    \bottomrule
    \multicolumn{5}{c}{(a): $\lambda_{sadv}$ for fully connected semantic-level adaptation module}
  \end{tabular}
  \end{subtable}
  
  \begin{subtable}{.49\linewidth}
  \centering
  \begin{tabular}{c|p{1.33cm}p{1.33cm}p{1.33cm}p{1.33cm}}
    \toprule
    \# City     & $\lambda=1$      & $\lambda=0.1$    & $\lambda=0.01$ & $\lambda=0.001$ \\
    \hline
    500                   & 59.76       & 59.46     & \textbf{60.16}     &59.67  \\
    \bottomrule
    \multicolumn{5}{c}{(b): $\lambda_{sadv}$ for CNN semantic-level adaptation module}
  \end{tabular}
  \end{subtable}
  \label{param}
  \vspace{-2mm}
\end{table}
\begin{table}
  \caption{SYNTHIA $\rightarrow$ Cityscapes: performance contributions of adaptation modules.}
  \vspace{-2mm}
  \footnotesize
  \label{SYNTHIAtable}
  \centering
  \begin{tabular}{c|cccc}
    \toprule
    \# City     & Oracle      & GA    & GA+CSA   & Improve\\
    \hline
    0                     & -           & 46.7     & -             & -\\
    50                    & 52.6       & 60.7     & 57.4         & +8.1\\
    100                   & 57.6       & 62.1     & 58.3         & +4.5\\
    200                   & 60.8       & 64.8     & 64.5         & +4.0\\
    500                   & 66.5       & 69.1     & 69.8         & +3.3\\
    1000                  & 70.7       & 71.8     & 73.0         & +2.3\\
    2975 (all)                 & 73.8       & 75.0     & 77.1         & +3.3\\
    \bottomrule
  \end{tabular}
  \label{syn_abl}
\vspace{-3mm}
\end{table}

\section{Experiments}
\begin{figure}
\centering
\includegraphics[width=.49\textwidth]{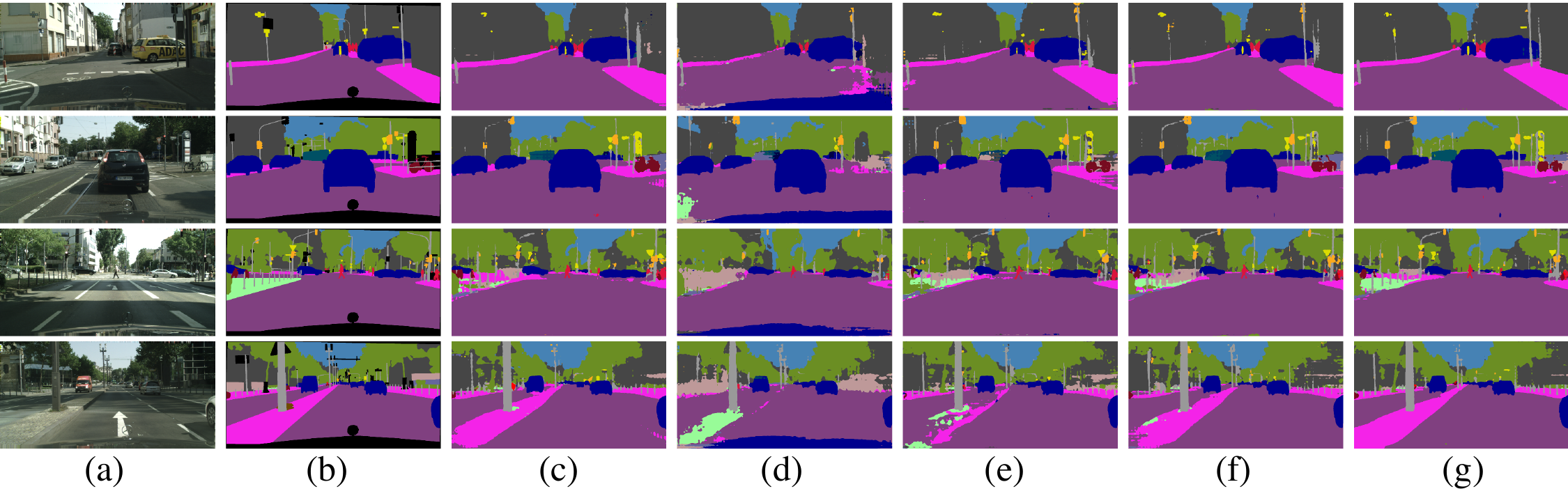}
\vspace{-5mm}
\caption{(a) image; (b) ground truth; (c) oracle model trained with the whole Cityscapes dataset; (d) unsupervised; (e) ours+200city; (f) ours+1000city; (g) ours+wholecity}
\label{figure4}
\vspace{-6mm}
\end{figure}

\vspace{-2mm}
We validate the effectiveness of our proposed method by transferring our model from a synthetic dataset (GTA5 \cite{gta5} or SYNTHIA \cite{synthia}) to a real-world image dataset Cityscapes \cite{Cityscapes}. The Cityscapes dataset contains 2975 images for training and 500 images for validation with 19-class fine-grained semantic annotations. following \cite{outputspace}, we first trained our model on the GTA5 dataset containing 19466 images and Cityscapes training set images and tested on the Cityscapes validation set for the whole 19 classes. The result is shown in Table \ref{gta_abl}. First, notice that the current state-of-the-art unsupervised model achieves 48.5 in mIoU \cite{bidir}. Our model can beat it by adding 50 Cityscapes images into the training process. This proves our argument that the model can have significant improvement by adding a few target domain information. Second, the contribution of $GA$ module disappears or is negative when the labeled Cityscapes images reach a number of 1000 or more compared to the oracle models. This is because the model with weak adaptation supervision overfits to the source domain so that it does not help much by adding relatively few more target images for the training process. However, the models $GA+FCSA$ and $GA+CSA$ both have on-par improvements if trained with over 50 Cityscapes labeled images. We argue that this is due to the strong adaptation supervision. Shown in Table \ref{param}, we observe that the $CSA$ and $FCSA$ structures are not very sensitive to the hyperparameters. We also provide some visualization results in Figure \ref{figure4}. Because $CSA$ is more laconic than $FCSA$, we only compare the model $GA+CSA$ with the other baseline models on transferring from Synthia dataset containing 9400 images to Cityscapes dataset. We compare the mIoU of 13 classes shared between SYNTHIA and Cityscapes \cite{outputspace} as shown in Table \ref{syn_abl}. The results can further support our arguments above.
\vspace{-1mm}

\vspace{-2mm}
\section{Conclusion}
\vspace{-1mm}
This paper proposes a semi-supervised learning framework to adapt the global feature and the semantic-level feature from the source domain to the target domain for the semantic segmentation task. As a result, with a few labeled target images, our model outperforms current state-of-the-art unsupervised models by a great margin. Our model can also beat the oracle model trained on the whole dataset from target domain by utilizing the synthetic data with the whole target domain labeled images without suffering from the prevalent problem of overfitting to the source domain. 
\vspace{-3mm}
\paragraph{Acknowledgment}
This work is supported by IBM-UIUC Center for Cognitive Computing Systems Research(C3SR).

{\small
\bibliographystyle{ieee_fullname}
\bibliography{egbib}

\begin{thebibliography}{10}\itemsep=-1pt

\bibitem{Nesterov}
Aleksandar Botev, Guy Lever, and David Barber.
\newblock Nesterov's accelerated gradient and momentum as approximations to
  regularised update descent.
\newblock In {\em IEEE IJCNN}, pages 1899--1903, 2017.

\bibitem{deeplabv1}
Liang-Chieh Chen, George Papandreou, Iasonas Kokkinos, Kevin Murphy, and Alan~L
  Yuille.
\newblock Semantic image segmentation with deep convolutional nets and fully
  connected crfs.
\newblock {\em arXiv preprint arXiv:1412.7062}, 2014.

\bibitem{deeplabv2}
Liang-Chieh Chen, George Papandreou, Iasonas Kokkinos, Kevin Murphy, and Alan~L
  Yuille.
\newblock Deeplab: Semantic image segmentation with deep convolutional nets,
  atrous convolution, and fully connected crfs.
\newblock {\em IEEE TPAMI}, 40(4):834--848, 2017.

\bibitem{deeplabv3}
Liang-Chieh Chen, George Papandreou, Florian Schroff, and Hartwig Adam.
\newblock Rethinking atrous convolution for semantic image segmentation.
\newblock {\em arXiv preprint arXiv:1706.05587}, 2017.

\bibitem{road}
Yuhua Chen, Wen Li, and Luc Van~Gool.
\newblock Road: Reality oriented adaptation for semantic segmentation of urban
  scenes.
\newblock In {\em IEEE CVPR}, 2018.

\bibitem{cheng2019spgnet}
Bowen Cheng, Liang-Chieh Chen, Yunchao Wei, Yukun Zhu, Zilong Huang, Jinjun
  Xiong, Thomas~S Huang, Wen-Mei Hwu, and Honghui Shi.
\newblock Spgnet: Semantic prediction guidance for scene parsing.
\newblock In {\em Proceedings of the IEEE International Conference on Computer
  Vision}, pages 5218--5228, 2019.

\bibitem{Cityscapes}
Marius Cordts, Mohamed Omran, Sebastian Ramos, Timo Rehfeld, Markus Enzweiler,
  Rodrigo Benenson, Uwe Franke, Stefan Roth, and Bernt Schiele.
\newblock The cityscapes dataset for semantic urban scene understanding.
\newblock In {\em IEEE CVPR}, 2016.

\bibitem{pascal}
Mark Everingham, S.~M.~Ali Eslami, Luc Van~Gool, Christopher K.~I. Williams,
  John Winn, and Andrew Zisserman.
\newblock The pascal visual object classes challenge: A retrospective.
\newblock {\em IJCV}, 111(1):98--136, 2015.

\bibitem{domainadapt}
Yaroslav Ganin, Evgeniya Ustinova, Hana Ajakan, Pascal Germain, Hugo
  Larochelle, Fran\c{c}ois Laviolette, Mario Marchand, and Victor Lempitsky.
\newblock Domain-adversarial training of neural networks.
\newblock {\em JMLR}, 17(1):2096--2030, 2016.

\bibitem{gan}
Ian~J. {Goodfellow}, Jean {Pouget-Abadie}, Mehdi {Mirza}, Bing {Xu}, David
  {Warde-Farley}, Sherjil {Ozair}, Aaron {Courville}, and Yoshua {Bengio}.
\newblock {Generative Adversarial Networks}.
\newblock {\em NIPS}, 2014.

\bibitem{hong}
Weixiang Hong, Zhenzhen Wang, Ming Yang, and Junsong Yuan.
\newblock Conditional generative adversarial network for structured domain
  adaptation.
\newblock In {\em IEEE CVPR}, 2018.

\bibitem{selferasing}
Qibin Hou, PengTao Jiang, Yunchao Wei, and Ming-Ming Cheng.
\newblock Self-erasing network for integral object attention.
\newblock In {\em NIPS}, 2018.

\bibitem{zilong}
Zilong Huang, Xinggang Wang, Jiasi Wang, Wenyu Liu, and Jingdong Wang.
\newblock Weakly-supervised semantic segmentation network with deep seeded
  region growing.
\newblock In {\em IEEE CVPR}, 2018.

\bibitem{huang2020alignseg}
Zilong Huang, Yunchao Wei, Xinggang Wang, Honghui Shi, Wenyu Liu, and Thomas~S
  Huang.
\newblock Alignseg: Feature-aligned segmentation networks.
\newblock {\em arXiv preprint arXiv:2003.00872}, 2020.

\bibitem{jiao2019geometry}
Jianbo Jiao, Yunchao Wei, Zequn Jie, Honghui Shi, Rynson~WH Lau, and Thomas~S
  Huang.
\newblock Geometry-aware distillation for indoor semantic segmentation.
\newblock In {\em Proceedings of the IEEE Conference on Computer Vision and
  Pattern Recognition}, pages 2869--2878, 2019.

\bibitem{adam}
Diederik~P. {Kingma} and Jimmy {Ba}.
\newblock {Adam: A Method for Stochastic Optimization}.
\newblock In {\em ICLR}, 2014.

\bibitem{leakyrelu}
Andrew L.~Maas, Awni Y~Hannun, and Andrew Ng.
\newblock Rectifier nonlinearities improve neural network acoustic models.
\newblock In {\em ICML}, 2013.

\bibitem{bidir}
Yunsheng Li, Lu Yuan, and Nuno Vasconcelos.
\newblock Bidirectional learning for domain adaptation of semantic
  segmentation.
\newblock In {\em The IEEE Conference on Computer Vision and Pattern
  Recognition (CVPR)}, June 2019.

\bibitem{qian2019weakly}
Rui Qian, Yunchao Wei, Honghui Shi, Jiachen Li, Jiaying Liu, and Thomas Huang.
\newblock Weakly supervised scene parsing with point-based distance metric
  learning.
\newblock In {\em Proceedings of the AAAI Conference on Artificial
  Intelligence}, volume~33, pages 8843--8850, 2019.

\bibitem{gta5}
Stephan~R. Richter, Vibhav Vineet, Stefan Roth, and Vladlen Koltun.
\newblock Playing for data: Ground truth from computer games.
\newblock In {\em ECCV}, 2016.

\bibitem{synthia}
German Ros, Laura Sellart, Joanna Materzynska, David Vazquez, and Antonio
  Lopez.
\newblock {The SYNTHIA Dataset}: A large collection of synthetic images for
  semantic segmentation of urban scenes.
\newblock In {\em IEEE CVPR}, 2016.

\bibitem{saleh}
Fatemeh Saleh, Sadegh Aliakbarian, Mathieu Salzmann, Lars Petersson, and
  Jose~M. Alvarez.
\newblock Effective use of synthetic data for urban scene semantic
  segmentation.
\newblock In {\em ECCV}, 2018.

\bibitem{dagan}
Swami Sankaranarayanan, Yogesh Balaji, Arpit Jain, Ser{-}Nam Lim, and Rama
  Chellappa.
\newblock Unsupervised domain adaptation for semantic segmentation with gans.
\newblock In {\em IEEE CVPR}, 2018.

\bibitem{outputspace}
Yi{-}Hsuan Tsai, Wei{-}Chih Hung, Samuel Schulter, Kihyuk Sohn, Ming{-}Hsuan
  Yang, and Manmohan Chandraker.
\newblock Learning to adapt structured output space for semantic segmentation.
\newblock In {\em IEEE CVPR}, 2018.

\bibitem{survey}
Mei Wang and Weihong Deng.
\newblock Deep visual domain adaptation: A survey.
\newblock {\em Neurocomputing}, 312:135--153, 2018.

\bibitem{wang2020differential}
Zhonghao Wang, Mo Yu, Yunchao Wei, Rogerior Feris, Jinjun Xiong, Wen mei Hwu,
  Thomas~S. Huang, and Honghui Shi.
\newblock Differential treatment for stuff and things: A simple unsupervised
  domain adaptation method for semantic segmentation.
\newblock {\em arXiv preprint arXiv:2003.08040}, 2020.

\bibitem{mining}
Yunchao Wei, Jiashi Feng, Xiaodan Liang, Ming-Ming Cheng, Yao Zhao, and
  Shuicheng Yan.
\newblock Object region mining with adversarial erasing: A simple
  classification to semantic segmentation approach.
\newblock In {\em IEEE CVPR}, 2017.

\bibitem{stc}
Yunchao Wei, Xiaodan Liang, Yunpeng Chen, Xiaohui Shen, Ming-Ming Cheng, Jiashi
  Feng, Yao Zhao, and Shuicheng Yan.
\newblock Stc: A simple to complex framework for weakly-supervised semantic
  segmentation.
\newblock {\em IEEE TPAMI}, 39(11):2314--2320, 2017.

\bibitem{revisit}
Yunchao Wei, Huaxin Xiao, Honghui Shi, Zequn Jie, Jiashi Feng, and Thomas~S.
  Huang.
\newblock Revisiting dilated convolution: A simple approach for weakly- and
  semi-supervised semantic segmentation.
\newblock In {\em IEEE CVPR}, 2018.

\bibitem{wu}
Zuxuan Wu, Xintong Han, Yen{-}Liang Lin, Mustafa~G{\"{o}}khan Uzunbas, Tom
  Goldstein, Ser{-}Nam Lim, and Larry~S. Davis.
\newblock {DCAN:} dual channel-wise alignment networks for unsupervised scene
  adaptation.
\newblock In {\em ECCV}, 2018.

\bibitem{fullyconv}
Yiheng Zhang, Zhaofan Qiu, Ting Yao, Dong Liu, and Tao Mei.
\newblock Fully convolutional adaptation networks for semantic segmentation.
\newblock In {\em IEEE CVPR}, 2018.

\bibitem{pspnet}
Hengshuang Zhao, Jianping Shi, Xiaojuan Qi, Xiaogang Wang, and Jiaya Jia.
\newblock Pyramid scene parsing network.
\newblock In {\em IEEE CVPR}, 2017.

\end{thebibliography}
}

\appendix
\appendixpage
\section{State-of-the-art comparison}
\begin{table*}
\caption{Results of adapting GTA5 to Cityscapes. The first four rows show the performance of the current state-of-the-art unsupervised algorithms. The following row shows the performance of our segmentation network trained on the whole Cityscapes dataset. The last six rows show the performance of our models trained with different number of Cityscapes labeled images.}
\scriptsize
\begin{center}
\begin{tabular}{ c|p{0.15cm}p{0.15cm}p{0.15cm}p{0.15cm}p{0.15cm}p{0.15cm}p{0.15cm}p{0.15cm}p{0.15cm}p{0.15cm}p{0.15cm}p{0.15cm}p{0.15cm}p{0.15cm}p{0.15cm}p{0.15cm}p{0.15cm}p{0.15cm}p{0.15cm}c }
 \hline
 \multicolumn{21}{c}{GTA5 $\rightarrow$ Cityscapes} \\
 \hline
Method & \begin{turn}{90}road\end{turn}
& \begin{turn}{90}sidewalk\end{turn}& \begin{turn}{90}building\end{turn}& \begin{turn}{90}wall\end{turn}& \begin{turn}{90}fence\end{turn}& \begin{turn}{90}pole\end{turn}& \begin{turn}{90}light\end{turn}& \begin{turn}{90}sign\end{turn}& \begin{turn}{90}vegetation\end{turn}& \begin{turn}{90}terrain\end{turn}& \begin{turn}{90}sky\end{turn}& \begin{turn}{90}person\end{turn}& \begin{turn}{90}rider\end{turn}& \begin{turn}{90}car\end{turn}& \begin{turn}{90}truck\end{turn}& \begin{turn}{90}bus\end{turn}& \begin{turn}{90}train\end{turn}& \begin{turn}{90}motorbike\end{turn}& \begin{turn}{90}bike\end{turn}& mIoU
\\
\hline
Wu et al.\cite{wu}&85.0&30.8&81.3&25.8&21.2&22.2&25.4&26.6&83.4&36.7&76.2&58.9&24.9&80.7&29.5&42.9&2.5&26.9&11.6&41.7
\\
Tsai et al.\cite{outputspace}&86.5&36.0&79.9&23.4&23.3&23.9&35.2&14.8&83.4&33.3&75.6&58.5&27.6&73.7&32.5&35.4&3.9&30.1&28.1&42.4
\\
Saleh et al.\cite{saleh}&79.8&29.3&77.8&24.2&21.6&6.9&23.5&44.2&80.5&38.0&76.2&52.7&22.2&83.0&32.3&41.3&27.0&19.3&27.7&42.5
\\
Hong et al.\cite{hong}&89.2&49.0&70.7&13.5&10.9&38.5&29.4&33.7&77.9&37.6&65.8&75.1&32.4&77.8&39.2&45.2&0.0&25.5&35.4&44.5
\\
\hline
oracle wholecity&96.7&75.7&88.3&46.0&41.7&42.6&47.9&62.7&88.8&53.5&90.6&69.1&49.7&91.6&\textbf{71.0}&73.6&45.3&\textbf{52.0}&65.5&65.9
\\
\hline
ours+50city&94.3&63.0&84.5&26.8&28.0&38.4&35.5&48.7&87.1&39.2&88.8&62.2&16.3&87.6&23.2&39.2&7.2&24.4&58.1&50.1
\\
ours+100city&96.0&71.7&85.9&27.9&27.6&42.8&44.7&55.9&87.7&46.9&89.0&66.0&36.4&88.4&28.9&21.4&11.4&38.0&63.2&54.2
\\
ours+200city&96.1&71.9&85.8&28.4&29.8&42.5&45.0&56.2&87.4&45.0&88.7&65.8&38.2&89.6&42.2&35.9&17.1&35.8&61.6&56.0
\\
ours+500city&96.2&72.7&87.6&35.1&31.7&46.6&46.9&62.7&88.7&49.6&90.5&69.2&42.7&91.1&52.6&60.9&9.6&43.1&65.6&60.2
\\
ours+1000city&96.8&76.3&88.5&30.5&41.7&46.5&51.3&64.3&89.1&54.2&91.0&70.7&48.7&91.6&59.9&68.0&40.8&48.0&67.0&64.5
\\
ours+2975city(all)&\textbf{97.3}&\textbf{79.3}&\textbf{89.8}&\textbf{47.4}&\textbf{49.7}&\textbf{48.9}&\textbf{52.9}&\textbf{67.4}&\textbf{89.7}&\textbf{56.3}&\textbf{91.9}&\textbf{72.2}&\textbf{53.1}&\textbf{92.6}&69.3&\textbf{78.4}&\textbf{58.0}&51.2&\textbf{68.2}&\textbf{69.1}
\\
\hline
\end{tabular}
\end{center}
\label{gtasota}
\vspace{-2mm}
\end{table*}
\begin{table*}[h!]
\caption{Results of adapting SYNTHIA to Cityscapes. The first three rows show the performance of the current state-of-the-art unsupervised algorithms. The following row shows the performance of our segmentation network trained on the whole Cityscapes dataset. The last six rows show the performance of our models trained with different number of Cityscapes labeled images.}
\scriptsize
\begin{center}
\begin{tabular}{ c|p{0.35cm}p{0.35cm}p{0.35cm}p{0.35cm}p{0.35cm}p{0.35cm}p{0.35cm}p{0.35cm}p{0.35cm}p{0.35cm}p{0.35cm}p{0.35cm}p{0.35cm}c }
 \hline
 \multicolumn{15}{c}{SYNTHIA $\rightarrow$ Cityscapes} \\
 \hline
Method & \begin{turn}{90}road\end{turn}
& \begin{turn}{90}sidewalk\end{turn}& \begin{turn}{90}building\end{turn}& \begin{turn}{90}light\end{turn}& \begin{turn}{90}sign\end{turn}& \begin{turn}{90}vegetation\end{turn}& \begin{turn}{90}sky\end{turn}& \begin{turn}{90}person\end{turn}& \begin{turn}{90}rider\end{turn}& \begin{turn}{90}car\end{turn}& \begin{turn}{90}bus\end{turn}& \begin{turn}{90}motorbike\end{turn}& \begin{turn}{90}bike\end{turn}& mIoU
\\
\hline
Wu et al.\cite{wu}&81.5&33.4&72.4&8.6&10.5&71.0&68.7&51.5&18.7&75.3&22.7&12.8&28.1&42.7
\\
Tsai et al.\cite{outputspace}&84.3&42.7&77.5&4.7&7.0&77.9&82.5&54.3&21.0&72.3&32.2&18.9&32.3&46.7
\\
Hong et al.\cite{hong}&85.0&25.8&73.5&19.5&21.3&67.4&69.4&68.5&25.0&76.5&41.6&17.9&29.5&47.8
\\
\hline
oracle wholecity&96.5&77.6&91.2&49.0&62.1&91.4&90.2&70.1&47.7&91.8&74.9&50.1&66.9&73.8
\\
\hline
ours+50city&94.1&63.9&87.6&18.1&37.1&87.5&89.7&64.6&37.0&87.4&38.6&23.2&59.6&60.7
\\
ours+100city&93.6&64.6&88.8&30.4&43.3&89.0&89.2&65.3&25.1&88.2&47.4&23.8&59.2&62.1
\\
ours+200city&95.2&71.2&89.1&32.9&46.4&89.1&90.3&67.0&31.6&89.4&42.3&32.9&64.8&64.8
\\
ours+500city&96.7&77.1&91.0&42.6&62.2&91.2&91.3&69.5&34.8&91.6&56.3&35.0&68.0&69.8
\\
ours+1000city&97.2&80.6&91.3&46.3&66.4&91.5&91.4&71.6&45.0&92.2&61.5&45.3&68.4&73.0
\\
ours+2975city(all)&\textbf{97.5}&\textbf{82.7}&\textbf{92.4}&\textbf{53.3}&\textbf{69.7}&\textbf{92.2}&\textbf{92.8}&\textbf{73.6}&\textbf{52.4}&\textbf{93.7}&\textbf{79.3}&\textbf{50.9}&\textbf{71.5}&\textbf{77.1}
\\
\hline
\end{tabular}
\end{center}
\vspace{-3mm}
\label{synsota}
\end{table*}
\end{document}